\documentclass[conference]{IEEEtran}
\IEEEoverridecommandlockouts
\usepackage{cite}
\usepackage{amsmath,amssymb,amsfonts}
\usepackage{graphicx}
\usepackage{textcomp}
\usepackage{xcolor}
\usepackage{booktabs}
\usepackage{tabularx}
\usepackage{booktabs}
\usepackage{colortbl} 
\usepackage{threeparttable}

\usepackage[linesnumbered, ruled, vlined]{algorithm2e}
\usepackage{hyperref}
\hypersetup{
    colorlinks=true,
    linkcolor=blue,
    filecolor=magenta,      
    urlcolor=cyan,
    pdftitle={Overleaf Example},
    pdfpagemode=FullScreen,
    }

\usepackage{float}
\usepackage{multicol, blindtext}

\def\BibTeX{{\rm B\kern-.05em{\sc i\kern-.025em b}\kern-.08em
    T\kern-.1667em\lower.7ex\hbox{E}\kern-.125emX}}

\begin{document}

\title{RNC: Efficient \underline{R}RAM-aware \underline{N}AS and \underline{C}ompilation for DNNs on Resource-Constrained Edge Devices

\author{
\IEEEauthorblockN{
    Kam Chi Loong\textsuperscript{1},
    Shihao Han\textsuperscript{1,3},
    Sishuo Liu\textsuperscript{1,4},
    Ning Lin\textsuperscript{1,2,3 $\dag$},
    Zhongrui Wang\textsuperscript{1,2,3 $\dag$}
}
\IEEEauthorblockA{
    \textsuperscript{1}Department of Electrical and Electronic Engineering, The University of Hong Kong, Hong Kong, China\\
    \textsuperscript{2}School of Microelectronics, Southern University of Science and Technology, Shenzhen, China \\
    \textsuperscript{3}ACCESS – AI Chip Center for Emerging Smart Systems, Hong Kong Science Park, Hong Kong, China\\
    \textsuperscript{4}School of Astronautics, Harbin Institute of Technology, Harbin, China\\
    \textsuperscript{$\dag$}Corresponding author: linning@hku.hk;wangzr@sustech.edu.cn
}}

}



\maketitle

\begin{abstract}

Computing-in-memory (CIM) is an emerging computing paradigm, offering noteworthy potential for accelerating neural networks with high parallelism, low latency, and energy efficiency compared to conventional von Neumann architectures. However, existing research has primarily focused on hardware architecture and network co-design for large-scale neural networks, without considering resource constraints. In this study, we aim to develop edge-friendly deep neural networks (DNNs) for accelerators based on resistive random-access memory (RRAM). To achieve this, we propose an edge compilation and resource-constrained RRAM-aware neural architecture search (NAS) framework to search for optimized neural networks meeting specific hardware constraints. 

Our compilation approach integrates layer partitioning, duplication, and network packing to maximize the utilization of computation units. The resulting network architecture can be optimized for either high accuracy or low latency using a one-shot neural network approach with Pareto optimality achieved through the Non-dominated Sorted Genetic Algorithm II (NSGA-II). 
The compilation of mobile-friendly networks, like Squeezenet and MobilenetV3 small can achieve over 80\% of utilization and over 6x speedup compared to ISAAC-like framework with different crossbar resources. The resulting model from NAS optimized for speed achieved 5x-30x speedup. The code for this paper is available at \url{https://github.com/ArChiiii/rram_nas_comp_pack}.

\end{abstract}


\section{Introduction}
The development of compact DNNs for smart edge devices poses significant challenges, particularly when inference must be performed locally without cloud computing support.
As edge devices often impose stringent hardware constraints on computing resources, battery capacity, system footprint (e.g., smart glasses, watches, ear buds) and latency (e.g., self-driving). Consequently, the search space for the optimal model diverges considerably from that of DNNs solely optimized for performance.
In addition, CMOS scaling is reaching its limit and the cost associated with huge data movement between processing and memory units is the bottleneck, which also known as ``von Neumann bottleneck".

To address these challenges, both DNN design and hardware need to be optimized.
For DNN design, Hardware-aware Neural Architecture Search (HW-NAS) has been used to address the challenges of diverse neural architectural design and optimization, alongside various hardware designs \cite{BenmezianeHadjer2021ACSo}.
For hardware, RRAM-based CIM accelerators show promising result for efficient DNN acceleration~\cite{PingChi2016PANP,ISAACShafieeAli2016IACN}.  It collocates data processing, memory, and storage, featuring high integration density, highly parallel matrix-vector multiplications and low power consumption.


However, existing CIM compilers and simulators \cite{ChenPai-Yu2018NACM, MNSIMZhuZhenhua2023M2AB, PIMCOMPSunXiaotian2023PAUC} lack support for mobile-friendly neural networks, limiting CIM applications on edge devices. Specifically, they do not support compilation and hardware metric estimation of depthwise convolution layers, which are crucial components in modern edge networks. These layers significantly reduce the number of parameters and computations. Without proper hardware simulation, it is impossible to accurately estimate the overall performance of the network on an RRAM-based accelerator.

Deploying neural networks on RRAM-based accelerators often results in low area utilization. The ISAAC-like \cite{ISAACShafieeAli2016IACN} mapping and benchmark results from NeuroSim \cite{ChenPai-Yu2018NACM} show inefficiency in RRAM crossbar utlization, especially for mobile-friendly networks. For Squeezenet \cite{iandola2016squeezenet} and MobilenetV3 \cite{howard2019searching}, the utilization of ISAAC-like mapping is 55\% and 45\% respectively. The utilization of NeuroSim for Squeezenet is 43\% and it is not compatible with mobile-friendly depthwise convolutional kernels. The low utilization results in additional latency and low energy efficiency. This issue has prompted extensive research into network design, hardware design, and compilation for RRAM-based accelerators.

Meanwhile, certain RRAM architectural designs lack generality. Recent studies~\cite{zhu2023pim,ICCAD2023Park} on hybrid architectural design of RRAM-based accelerator reported satisfactory performance by combining crossbars with different sizes in a single CIM architecture accelerator. However, these designs are tailored for particular neural networks, rather than a general hardware solution for edge model deployement. Only limited works~\cite{YuanZhihang2021Nnna,SunHanbo2022GECo} emphasize the importance of RRAM-aware NAS, considering the specific properties of RRAM crossbars and peripheral circuit, including array size and quantization errors. 








To optimally tackle the aforementioned challenges, we introduce the first RRAM-aware NAS and compilation framework, termed \textit{RNC}. The key contributions of this paper are summarized as follows:
\begin{itemize}
\item \textbf{The first edge compilation for mobile DNNs:} \textit{RNC} compiles network with novel layer partition for edge-friendly deployment on RRAM crossbars. Efficient duplication enables high parallelism when extra crossbars are available.

\item \textbf{A novel network packing method for high RRAM crossbar utilization:} 
\textit{RNC} converts weight mapping to a bin-packing problem (BPP) and develops network packing with heuristics to optimize hardware metrics, area utilization, and latency for edge networks. The packing can achieve over 80\% of crossbar utilization.


\item \textbf{RRAM-aware NAS for efficient DNNs on edge:}
\textit{RNC} introduces RRAM-aware NAS using Non-Dominated Sorting Genetic Algorithm II (NSGA-II) with hardware metrics feedback to search for optimized neural architecture with Pareto optimality. The speed optimized model demonstrate 5.8x-7.5x improvement and the accuracy optimized model shows 2\% improvement.
\end{itemize}

The rest of paper is organised as following: Section \ref{sec:preliminaries} introduce RRAM, CIM compilers and simulator, and NAS. Section \ref{sec:overview} presents the RNC in details. Section \ref{sec:evaluation} presents the evaluations of compilation and NAS. Section V concludes this paper.

\begin{figure}[!tb]
    \centering
    \includegraphics[width=0.4\textwidth]{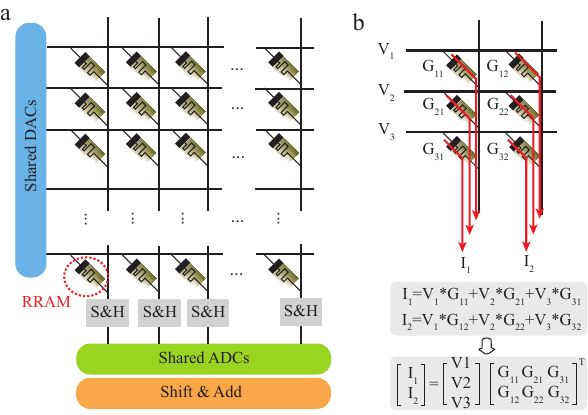}
         \vspace{-6pt}  
    \caption{(a) RRAM crossbar array structure. (b) Ohm's Law and Kirchoff’s Law in for analog MAC operations with a RRAM crossbar.}
         \vspace{-12pt}  
    \label{fig:rram_preli}
\end{figure}

\section{Preliminaries} \label{sec:preliminaries}


\subsection{RRAM based CIM Accelerators}
Over the past decade, numerous hardware DNN accelerators have emerged with the aim of enhancing both energy and area efficiency. Conventional von Neumann architecture accelerators \cite{TPU2017ISCA,isscc_2016_chen_eyeriss} are plagued by significant energy and time overheads due to the constant transfer of data between the main memory and computing components. To address the von Neumann bottleneck, M accelerators, which combine memory and processing units, are introduced. 
RAM is a compact, cost-effective, and mature nonvolatile memory technology that enables analog CIM. By utilizing Ohm's law and Kirchhoff's law, RRAM crossbars efficiently carry out multiply–accumulate (MAC) operations with a high degree of parallelism.

In an RRAM-based CIM macro, RRAM cells are organized in a crossbar array, as shown in Fig. \ref{fig:rram_preli}. Weight matrices are stored as resistive states of the RRAM cells. Input vectors are converted into row biasing voltages using digital-analog converters (DAC). Each resulting current vector is the sum of the currents flowing through cells of conductance (\(G_{xy}\)) biased by applied voltages (\(V_y\)) along a column, using Ohm's law for multiplication and Kirchhoff's Law for summation. This configuration enables the entire array to perform analog MAC operations of DNNs in parallel without accessing memory to fetch weights.
Subsequently, a sample-and-hold (S\&H) circuit captures the bitline current and directs it to a shared ADC and Shift \& Add, thereby completing the multi-bit matrix multiplication in analog domain.


\subsection{CIM Compilers and Simulator}
The CIM compiler is a specialized tool developed to deploy DNNs on the CIM architecture hardware. PIMCOMP \cite{PIMCOMPSunXiaotian2023PAUC} is a comprehensive compilation system that supports node partitioning, weight replication, and dataflow scheduling in CIM system. MNSIM 2.0 \cite{MNSIMZhuZhenhua2023M2AB} is a behavior-level modeling tool that offers a hierarchical solution for deploying models on digital and analog CIM systems. This software optimizes CIM systems at various levels, aiding in performance assessment, design exploration, and algorithm and hardware optimization. Typically, a simulator assesses the hardware performance. Besides, NeuroSim \cite{ChenPai-Yu2018NACM} is a simulator that operates at the circuit level and calculates hardware performance metrics for CIM accelerators, with a specific focus on area, latency, etc. These tools collectively enhance the design and implementation of CIM architectures.

In this work, we follow the universal compilation steps introduced in PIMCOMP \cite{PIMCOMPSunXiaotian2023PAUC}. In layer/node partition, we introduce a new technique for depthwise convolution, which is not supported in previous work. We also formulate the weight duplication as a constrained optimization that can be solved efficiently. While previous works map individual weight matrices to individual crossbars, we map weight matrices in granularity of crossbar and improve the utilization of hardware resource. This allows us to deploy machine learning models of more parameters and better performance on resource-limited edges. Meanwhile, we allow strict crossbar resource constraints that is frequent in edge applications. At the same time, we also leverage the simulation in MNSIM 2.0 \cite{MNSIMZhuZhenhua2023M2AB} to evaluate the performance within the crossbar of our compilation. 

\begin{table}[!tb]
\centering
\caption{Features of CIM compilers and simulators.}
\vspace{-6pt}  
\label{table:cim_compiler}

\begin{threeparttable}

\scalebox{0.7}{
\begin{tabular}{c|cccc}
\toprule
 & \textbf{NeuroSim}~\cite{ChenPai-Yu2018NACM} & \textbf{MNSIM2.0} ~\cite{MNSIMZhuZhenhua2023M2AB} & \textbf{PIMCOMP}~\cite{PIMCOMPSunXiaotian2023PAUC} & \textbf{RNC (This work)} \\
\midrule
Accuracy Sim. & High & High & Low & Medium \\
\cellcolor[HTML]{EFEFEF} Layer Partition & \cellcolor[HTML]{EFEFEF} Yes & \cellcolor[HTML]{EFEFEF} Yes & \cellcolor[HTML]{EFEFEF} Yes & \cellcolor[HTML]{EFEFEF} Yes \\
Weight Replication & Yes & No & Yes & Yes \\
\cellcolor[HTML]{EFEFEF} Weight Mapping & \cellcolor[HTML]{EFEFEF} PE & \cellcolor[HTML]{EFEFEF} PE & \cellcolor[HTML]{EFEFEF} PE & \cellcolor[HTML]{EFEFEF} Crossbar \\
Constraint Sup.& Low & Medium & Medium & High \\
\cellcolor[HTML]{EFEFEF} Operation Sup.& \cellcolor[HTML]{EFEFEF} FC/Conv & \cellcolor[HTML]{EFEFEF} FC/Conv & \cellcolor[HTML]{EFEFEF} FC/Conv & \cellcolor[HTML]{EFEFEF} FC/Conv/DW Conv {\color{red} \tnote{1}} \\
\bottomrule
\end{tabular}
}

\begin{tablenotes}
\footnotesize
\item[{\color{red} 1}] FC: Fully Connected Layer; Conv: Convolutional Layer;\\ DW Conv: Depthwise Convolutional Layer.
\end{tablenotes}
\vspace{-12pt}  
\end{threeparttable}
\end{table}

\begin{figure*}[ht]
    \centering
    \includegraphics[width=0.6\textwidth]{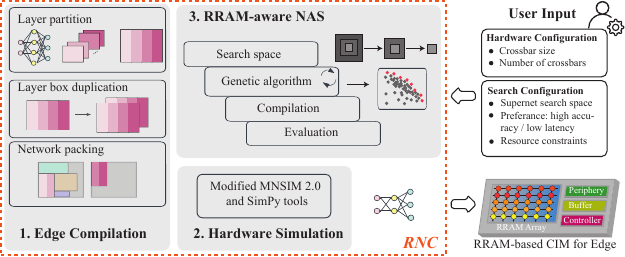}
         \vspace{-6pt}  
    \caption{Overview of efficient RRAM-aware NAS and compilation for DNNs on resource-constrained edge devices.}
    \label{fig:overview}
\vspace{-18pt}
\end{figure*}

\subsection{Neural Architecture Search}
NAS is a swiftly expanding subfield of machine learning that concentrates on automating the process of designing neural network structures. NAS  systematically explores a predetermined search space in order to uncover desire models. Conventional NAS utilizes different search tactics, including evolutionary algorithms \cite{9533986}, reinforcement learning \cite{BarretZoph2017NASw}, and gradient-based methods \cite{liu2019darts} to explore the search space and find the best model.

HW-NAS~\cite{benmeziane2021comprehensive} integrates hardware restrictions and performance data into the search process. HW-NAS search for models by taking into account hardware performance such as latency, power consumption, and memory utilization. This co-design methodology guarantees that the generated models are not only precise but also efficient, scalable, and cost-effective for implementation on many hardware platforms, ranging from edge devices to high-performance servers.

\section{Overview of RNC} \label{sec:overview}

The overview of RNC is depicted in Fig. \ref{fig:overview}. User can provide hardware configuration and network specification for edge compilation and performance simulation. With a pre-defined network search space, RRAM-aware NAS identifies an optimized network that meets specified constraints and preferences. 


\begin{figure}[!tb]
    \centering
    \includegraphics[width=0.45\textwidth]{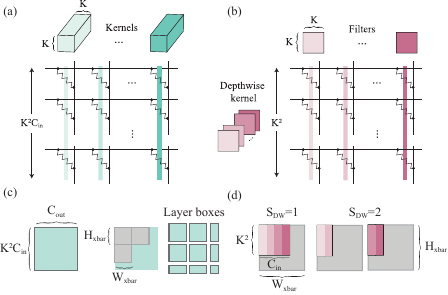}
         \vspace{-12pt}  
    \caption{Layer partition method. (a) Kernels flatten to a 2D weight matrix for standard convolution. (b) Filters flatten to a 2D weight matrix for depthwise convolution. (c) Partition for an oversize weight matrix to multiple layer boxes. (d) Effect of depth split factor.}
         \vspace{-16pt}  
    \label{fig:comp_partition}
\end{figure}

\subsection{Edge Compilation} \label{section_compile}
We optimize the area utilization by casting it as a classic BPP. The kernels of each convolution and dense layer are flattened into a 2D layer box which is then packed inside the crossbar arrays, analogous to containers in the original BPP. We augment BPP by layer partition, layer box duplication, in addition to packing.

\subsubsection{Layer Partition}
Layer partition divides and creates a layer box from the weight matrix of each layer. It also ensures oversized network layer can be mapped onto crossbar arrays by dividing large kernels into layer boxes of manageable sizes. Additionally, it handles depthwise separable convolution layers to reduce the processing cycle with heuristic packing methods.

As in previous work~\cite{MNSIMZhuZhenhua2023M2AB,PIMCOMPSunXiaotian2023PAUC}, a kernel is flattened into a 1D array, usually a convolution layer consists of multiple kernels and result in a 2D layer box with a height of \( k_w \times k_h \times C_{in}\) and a width \( C_{out} \), which is shown in Fig. \ref{fig:comp_partition}(a). Due to the crossbar size limitation, the layer box may exceed the container size, necessitating partitioning based on crossbar dimensions. The layer box can result in four possible sizes, including same size as the crossbar, same width but different height, same height but different width, and different width and height, Fig.~\ref{fig:comp_partition}(c) shows the details. The latency cycle of dense layer box and normal convolution layer box are 1 and \( H_{out} \times W_{out} \), which are the height and width of the output feature map. 



Unlike normal convolution layers, depthwise convolution layers are inherently crossbar-unfriendly.
Instead of convolution across all input channels, in depthwise convolution, each input channel is convolved with a different kernel. When combined with pointwise convolution, this results in depthwise separable convolution, which significantly reduces the number of computations. Depthwise layer is first flattened with its filters, shown in \ref{fig:comp_partition}(b), resulting in 2D layer box with height of \( k_w \times k_h \) and width of \(C_{in}\). However the cycle would be \( H_{out} \times W_{out} \times C_{in} \) as each kernel filter does not share the input. Therefore, we propose division of the depthwise layer box by a split factor \( S_{DW} \) along the width, showing in Fig. \ref{fig:comp_partition}(d). The new spitted layer boxes would require reduced cycles of \( H_{out} \times W_{out} \times C_{in}/S_{DW} \). The depth split factor \( S_{DW} \) is crucial to balance number of layer boxes and the cycles within the constraints.

\subsubsection{Layer Box Duplication}
Layer box duplication addresses low utility of crossbar resources as well as increases parallelism and throughput of the system \cite{PIMCOMPSunXiaotian2023PAUC}. To balance the latency of all layers in the pipeline, we duplicate time-consuming layers to minimize overall latency. This duplication can be formulated as a constrained optimization problem as follows,
\begin{align}
\text{Minimize: } & \quad \sum \frac{C_i}{x_i}  \\
\text{Subject to: } & \quad \sum x_i A_i \leq A_{xbar} \notag,
\end{align}
where \(C_i\) is latency of layer i and \(x_i\) is the number of copies of layer i, which equals to minimising the total latency after duplication subject to total crossbar capacity constraint. This is a non-linear problem and we try to transform it into linear problem by estimate the number of inference samples per latency as \(K_i\) where \(K_i \approx \frac{1}{C_i}\). The problem can be reformulated as follows,
\begin{align}
\text{Maximize: } & \quad {K_t}{x_t}  \\
\text{Subject to: } & \quad \sum x_i A_i \leq A_{xbar} \notag \\
\text{} & \quad {K_i}{x_i} \geq {K_t}{x_t} \quad \forall i \in \{0, 1, 2, ..., i\} \notag,
\end{align}
where \(K_t\) is number of inference samples per latency of the most time-consuming layer \(x_t\). We optimize the output of layer \(x_t\) with the  constraint of all other layers to be equal or better than it. The area constraint remains the same. The above optimization can be solved efficiently by linear programming. The optimized result determines the copies of each layer box.


 
 

  

            
                


                
            


\subsubsection{Network Packing}

Network packing is a variant of classic bin packing, which is a NP-hard combinatorial optimization problem. In this context, network packing involves arranging layer boxes of various sizes into a fixed number of crossbar containers without rotation. Maintaining the orientation of layer boxes is crucial to ensure correct data flow as crossbar inputs/outputs are with different box edges.

Network packing enhances the utilization of the crossbar arrays by flexibly mapping weight matrices with the granularity of crossbar arrays. This allows to accommodate a larger model on edge devices, resulting in 
improved network performance. The benefit is even more pronounced when standard convolution layers are replaced with depthwise separable convolutions. These convolutions, which combine depthwise and pointwise layers, drastically reduce the number of parameters, which is also proportional to areas of the layer boxes. Take a (3,3,32) convolution kernel with 3 input channels as an example. Normal convolution would have \(3 \times3 \times3 \times32=864\) parameters while depthwise separable convolution would only require \(3 \times3 \times3 + 32=59\) parameters. Thus, combination of packing and depthwise separable convolution can significantly improve the model capability without increasing RRAM crossbar capacity. 

To achieve effective and efficient packing, we utilize heuristic packing methods due to their simplicity and scalability. Specifically, we adopt Empty Maximal Space (EMS)\cite{cdi_springer_books_10_1007_978_3_319_55792_2_10}  with Distance to the Front-Top-Right Corner (DFTRC)\cite{GonçalvesJoséFernando2013Abrk} distance as our placement rule. 

\begin{figure}[!t]
    \centering
    \includegraphics[width=0.3\textwidth]{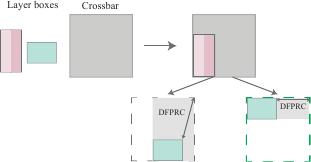}
         \vspace{-6pt}  
    \caption{Placement rule with EMS and DFTRC. Placement with less DFTRC is selected.}
         \vspace{-10pt}  
    \label{fig:packing}
\end{figure}

EMS refers to the largest available space within the container that remains after placement of layer box. It is the potential placement for next layer box. Fig. \ref{fig:packing} shows the EMS in gray after the placement of a layer box. The target space is selected by the shortest distance between the top right corner of the layer box and container, which is a 2D variant of DFTRC.
The algorithm details are shown in Algorithm \ref{alg:packing_ems}.

\begin{algorithm}[t]
	\caption{RNC Packing with EMS and DFTRC}
	\label{alg:packing_ems}
	\KwIn{Set of layer boxes $B$, Set of containers $C$ }
	\BlankLine
 
	Initialize empty spaces of $C$
        
        $E$ = set of empty space 
        
        \ForEach{$b \in B$}{
        
            Select available EMS from $E$ using DFTRC heuristic and layer collision constraint
            
            \If{Found}{
                Place $b$ in the selected EMS
                
                Update $E$ by:

                - Removing the selected EMS

                - Adding new EMSs resulting from the placement

                Remove EMSs inscribed by other EMSs
                
            }\Else{
            
                Continue to the next container
                
            }      
            
        }
        
        \KwOut{Packing configuration}  
        
\end{algorithm}

Additionally, we implement a critical heuristic that prohibits packing layer boxes from the same layer or from the immediately consecutive layer into the same crossbar. Since we assume only one layer box is activated for computing per crossbar at a time, packing layer boxes from the same layer together can lead to structural hazards and result in stalls. Similarly, packing layer boxes from adjacent layers can cause data hazards. This heuristic mitigates the impact of structural hazards and data hazards. As the number of inference samples grows, stalls become inevitable. At the same time, the heuristic facilitates parallel computation by distributing duplicated layer boxes across different RRAM crossbars.

By adopting these strategies, network packing of RNC ensures that the available crossbar resources are used optimally, thereby improving the performance of neural networks deployed on edge devices.




\subsection{Hardware Simulation}
To assess latency, we employ a cycle-accurate simulator from modified MNSIM 2.0~\cite{MNSIMZhuZhenhua2023M2AB} integrated with SimPy\cite{simpy} to accurately capture the active periods and idle periods of individual layer boxes deployed on crossbar arrays. We utilize MNSIM 2.0 for time evaluation and manage the scheduling with SimPy.

The simulation activates layer boxes within RRAM crossbars according to model layer sequence. It considers the parallelism of duplicated layer boxes and provides detailed analysis of latency impacts arising from resource conflicts and data dependency. To evaluate the total system latency, we combine the execution times of each layer. This aids NAS in determining the most efficient network configuration that minimize latency on edge scenarios like self-driving where response time is of great importance.



\begin{figure}[!tb]
    \centering
    \includegraphics[width=0.45\textwidth]{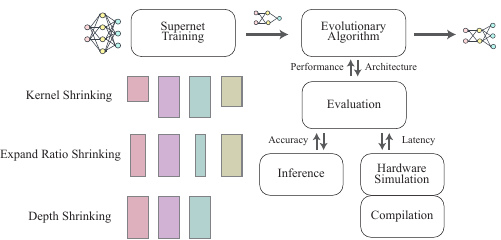}
    \vspace{-6pt}
    \caption{RRAM-aware Neural Architecture Search. }
    \label{fig:nas_flow}
\vspace{-14pt}
\end{figure}

\subsection{RRAM-aware NAS}
\subsubsection{Training of One-shot Supernet}
RNC is a one-shot NAS framework that defines an overparameterized network as supernet that encompasses weights of all child networks. The supernet is trained once and NAS samples child networks from the supernet for evaluation without retraining\cite{CaiHan2020OTON}. This weight sharing approach can significantly reduce training costs compared to traditional NAS and able to provide a versatile design space for various hardware architectures. Trained supernets can be reused for different NAS search objectives.

The supernet is organized into blocks according to the base model. It supports elastic blocks and elastic layers. Elastic blocks can contain varying numbers of layers in the block, while elastic layers comprise elastic kernels that accommodate various kernel sizes and channels. Training the supernet is not trivial and we follow technique called progressive shrinking \cite{CaiHan2020OTON} to train the entire model and fine-tune the model with shrank attribute in each dimension. The details can be found in Algorithm \ref{alg:progressive_strinking}.

\begin{algorithm}[ht]
	\caption{Training with progressive shrinking}
	\label{alg:progressive_strinking}
	\KwIn{Kernel size options $K_s$; Channel size options $W$; 
 Depth options $D$ }
	\BlankLine
 
	Define a supernet with  $K_s$,$W$,$D$
 
        Train the network with $Max(K_s)$,$Max(W)$,$Max(D)$

        $\mathcal{D} = \{K_s, W, D\}$
        
        \ForEach{$d \in \mathcal{D}$}{

            Sort $d$ in descending order
            
            \ForEach{option in $d$ }{
            
                Fine-tune the model with $option$
            }
        }
     \KwOut{Well-trained Supernet}  
\end{algorithm}

\subsubsection{Search of Specialised Network}

The primary goal of RNC is to identify the most efficient neural network model under restrictive hardware constraints, such as a limited number of crossbar arrays or stringent latency. These constraints are particularly important in edge computing scenarios where resources are limited. Typically, larger networks provide better model performance in the cost of higher latency and resource consumption. Hense, RNC employs a multi-objective optimization strategy to balance these trade-offs, resulting in a Pareto front. Pareto front contains a set of non-dominated child networks. We can select a model that is optimized for either accuracy or latency. The metrics are obtained after the edge compilation and evaluation. The overall flow is presented in Fig. \ref{fig:nas_flow}.


\begin{algorithm}[t]
	\caption{One shot RRAM-NAS with NSGA-II}
	\label{alg:rramnas}
	\KwIn{Hardware constraint $C_{HW}$; Search Preference $P$; 
 Population $N$ }
	\BlankLine
 
	Train a supernet $\mathcal{T}$
 
    Initialise population ${P(t)}$ with subnet $t$
    
    \ForEach{generation}{
        Evaluate model performance of ${P(t)}$
        
        Evaluate hardware metrics of ${P(t)}$ with $C_{HW}$

        Assign rank $R$ based on Pareto
        
        Calculate crowding distance $D$
        
        Add non-dominated solution to next generation $g$ with higher $R$ and $D$ until $N$ individuals
        
        ${P(t)}$ = next generation with crossover and mutation 
    }
 \KwOut{Optimised neural architecture with $P$ from highest rank Pareto set}

\end{algorithm}

We search for network in supernet with NGDA-II \cite{LuZhichao2019NNAS} \cite{DebK.2002Afae} and introduce RRAM-specific metric as one objective together with model performance (e.g., classification accuracy). NSGA-II is well-suited for multi-objective optimization as it effectively handles trade-offs between competing objectives. The detailed algorithm is listed in Algorithm \ref{alg:rramnas}.





\section{Evaluations} \label{sec:evaluation}
\subsection{Evaluation Setup}
\subsubsection{Edge Compilation}
We performed edge compilation with Resnet18 \cite{he2015deep}, Squeezenet \cite{iandola2016squeezenet} and MobilenetV3 small\cite{howard2019searching}. They are representative convolution networks for edge image classification. Resnet18 is relatively deeper and larger network, which typically offers better accuracy due to its residual connections. Squeezenet is designed as a lightweight network, achieving high accuracy with fewer parameters by using fire modules, which are a combination of squeeze and expand layers. MobilenetV3 small is optimized for mobile and edge devices, utilizing depthwise separable convolutions and efficient operations such as squeeze-and-excitation (SE) blocks, resulting in fewer parameters and lower computational cost, making it ideal for resource-constrained environments.

The evaluation starts with compilation of those networks with the minimum number of crossbars and compares their performance with ISAAC-like \cite{ISAACShafieeAli2016IACN} mapping, which only maps one layer per crossbar. ISAAC-like mapping ensures there is no structural hazard as the layer can be executed in full parallel, resulting in large number of crossbars and low utilization. We further show the contribution of different steps in the compilation, including the depthwise layer box splitting with different depth split factor $S_{DW}$ and the layer box duplication when more crossbars are available. The search objective values the number of crossbars, crossbar utilization, and de the  nt numbers of processing samples. All the compilation is done on crossbars wi128x128 th size oan ideal peripheral circuit and the lat.eTf the core connection, data transmission and bandwidth are neglected. All the operations are executed layer by layer. 

\subsubsection{RRAM-aware NAS}

We performed one-shot RRAM-aware NAS for MobilenetV3 small for scaled CIFAR-10 dataset \cite{cifar10}. Detailed search space parameters are shown in Table \ref{table:search_space}. The search space is applied to the bottleneck blocks. These bottleneck blocks are organized into 4 groups, each representing a set of layers where the spatial or channel dimensions of the feature maps remain constant within the group, but change between groups. Specifically, the input and output dimensions differ as we transition between groups due to downsampling or increasing the number of channels. The supernet is trained by progressive shrinking \cite{CaiHan2020OTON}. As MobilenetV3 small is designed to target large images, the spatial dimension of intermediate feature map would lose important information with original image dimension of CIFAR-10, so we scaled up CIFAR-10 images from 32x32 to 64x64.

The evaluation is conducted by comparing four different methods. ISAAC-like mapping is the first one, which is the baseline. The second is the compilation result that uses the same resource constraints as the baseline but applies optimized mapping techniques for improved performance. The third and fourth are the results of an RRAM-aware NAS that is optimized for accuracy and speed, respectively. They are compared in terms of accuracy, speed, and utilization.

\begin{table}[h]
\centering
\def\arraystretch{1.5}%
\vspace{-10pt}
\caption{Supernet search space.}
     \vspace{-6pt}  
\begin{tabular}{c||c}
\hline
\textbf{Kernel Size Set}  & \{3, 5, 7\} \\ \hline
\textbf{Expand Ratio Set} & \{3, 4, 6\} \\ \hline
\textbf{Depth Set}        & \{2, 3, 4\} \\ \hline
\end{tabular}
\label{table:search_space}
\vspace{-6pt}
\end{table}

We search the network with given hardware constraint. The hardware constraint would be the number of crossbars in compilation. The best model is selected from a Pareto set with either accuracy or latency optimized. The latency is normalized with respect to that of the original MobilenetV3 small model using ISAAC-like mapping. Additionally, we examine the compilation performance with different number of crossbars.


For NSGA-II, the population of the algorithm is 50 and producing maximum of 100 generations with 0.25 mutation probability.






\subsection{Evaluation Results}

\begin{figure}[!t]
    \centering
    \includegraphics[width=0.5\textwidth]{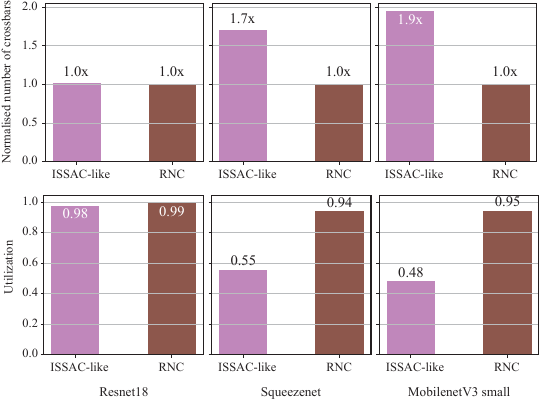}
    \vspace{-20pt}
    \caption{Normalized number of crossbars and utilization of Resnet18 \cite{he2015deep}, Squeezenet \cite{iandola2016squeezenet} and MobilenetV3 small \cite{howard2019searching} on minimum number of crossbars of size 128x128 under ISAAC-like mapping and RNC compilation.}
    \label{fig:compilation_result}
    \vspace{-8pt}
\end{figure}

\begin{figure}[!t]
    \centering
    \includegraphics[width=0.3\textwidth]{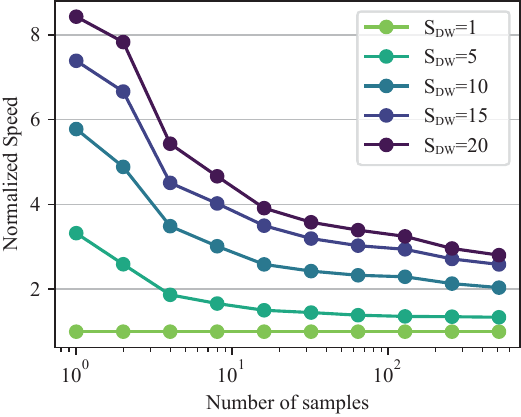}
    \vspace{-6pt}
    \caption{Normalized speed of MobilenetV3 small with different depth split $S_{DW}$ and minimum number of crossbars of size 128x128.}
    \label{fig:compilation_depth_split}
\vspace{-10pt}
\end{figure}

\begin{figure}[!t]
    \centering
    \includegraphics[width=0.5\textwidth]{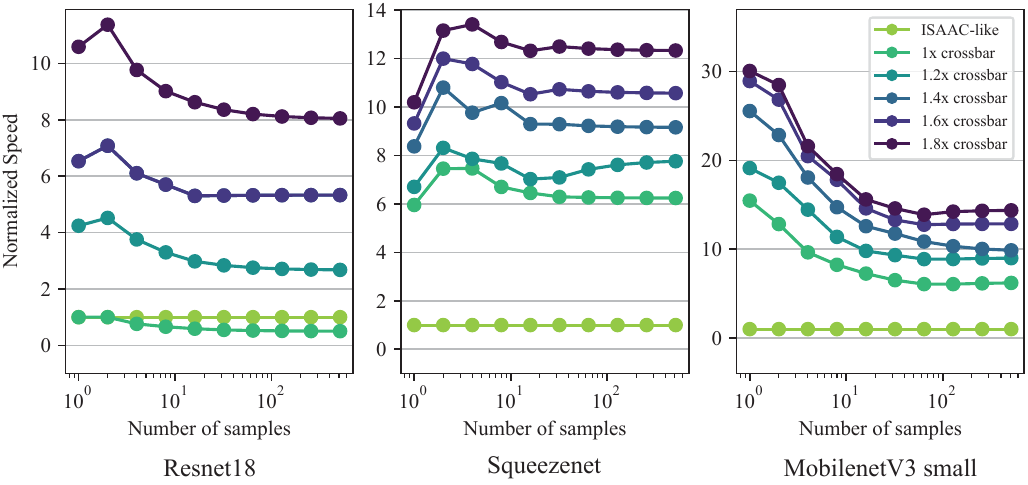}
    \vspace{-20pt}
    \caption{Speed of different networks under the edge compilation with different number of crossbars. (\(S_{DW}=1\) for MobilenetV3 small.)}
    \label{fig:compilation_duplication_result}
\vspace{-16pt}
\end{figure}

\begin{figure*}[!t]
    \centering
    \includegraphics[width=0.85\textwidth]{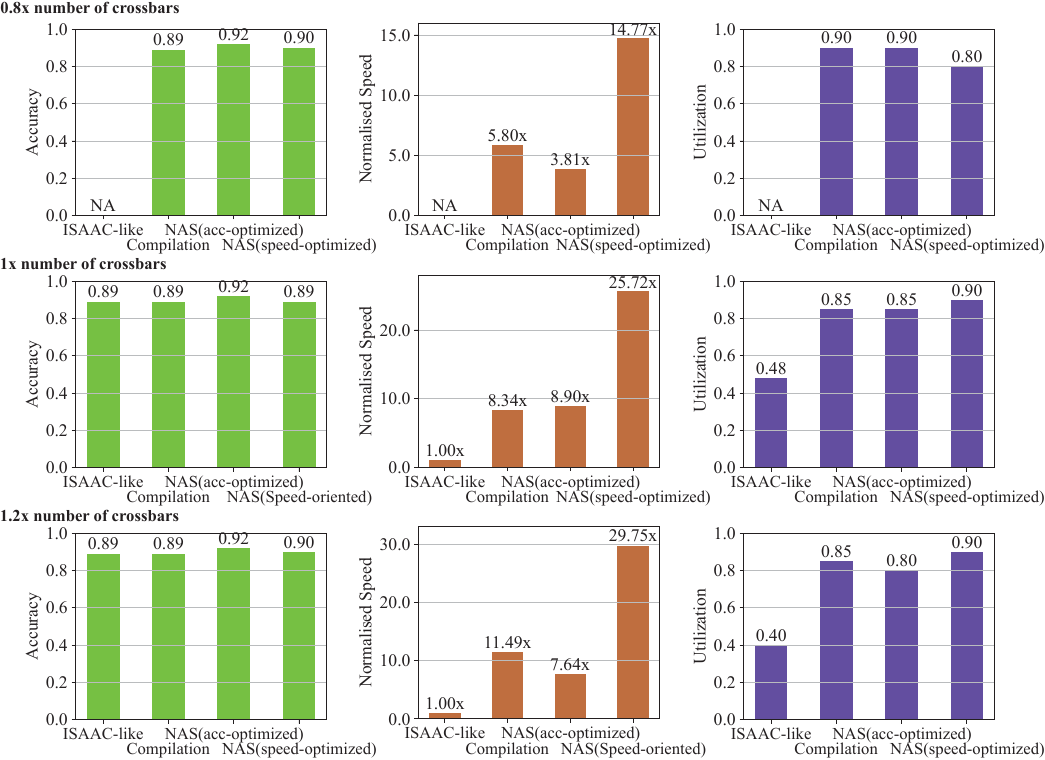}
    \vspace{-6pt}
    \caption{Analysis on accuracy, latency and utilization of ISAAC-like structure, edge compilation result and optimized result from RRAM-aware NAS.}
    \label{fig:nas_result}
\vspace{-18pt}
\end{figure*}

\subsubsection{Edge Compilation}

After layer partition and packing, the result of packing with minimum number crossbars (i.e. no duplication) is illustrated in Fig. \ref{fig:compilation_result}, showing the improvement of utilization and reduce of crossbars, as the layer boxes are efficiently grouped together. The compilation can achieve almost 2x utilization improvement of the lightweight networks and reduce over 40\% of the crossbars required. The packing shows almost no effect on Resnet18 as most layer boxes exceed the size of crossbar. Therefore, a layer box in Resnet18 can take up an entire crossbar array, which makes the result similar to baseline.

Depthwise splitting is uniquely applicable to MobilenetV3 small, since the other two networks do not have depthwise layers. The normalized speed of MobilenetV3 small (with processing sample sizes from 1 to 1024 in powers of 2) is presented in Fig. \ref{fig:compilation_depth_split}. The baseline corresponds to the case $S_{DW}=1$, with no depthwise splitting. For single sample inference, the splitting can achieve up to 8x speedup with $S_{DW}=20$. As the number of inference samples increase, the effectiveness diminishes due to overheads, while $S_{DW}=10$ can still have 2x improvement for sample sizes between 128 and 512. 

Fig. \ref{fig:compilation_duplication_result} presents the speedup achieved with varying number of crossbars, using the ISAAC-like mapping as the baseline. In ISAAC-like mapping, each layer is mapped on an individual crossbar, leaving substantial room for optimization due to underutilization of crossbar resources.The ``1x crossbar" configuration in our results represents the same number of crossbars as the ISAAC-like baseline but with our optimized compilation strategy.

For ResNet18, we observe performance degradation in the 1x crossbar configuration due to structural conflicts caused by layer duplication. The large layer size often requires a single layer to span multiple crossbars, increasing the likelihood of conflicts when duplicated layers are mapped to the same crossbars. However, as the number of crossbars increases, these conflicts are mitigated, enabling more efficient layer mapping and resulting in substantial speedup gains. For example, a configuration with 1.8x crossbars achieves over an 8x speedup compared to the baseline.

In contrast, lightweight networks such as Squeezenet and MobileNetV3 demonstrate significant speedups even with the 1x crossbar setting. For these networks, the optimized compilation results in over 6x speedup for Squeezenet and 6x-15x for MobileNetV3 when compared to the ISAAC-like baseline. This substantial improvement is primarily because the smaller layers in these networks can be duplicated multiple times without causing structural conflicts, leading to better parallelism and crossbar utilization.




In Fig. \ref{fig:nas_result}, our compilation outperforms baseline by over 8.3x speedup while archiving around 2x of utilization with the same number of crossbars. When the number of crossbars is restricted to 80\%, the compilation can successfully map all weights onto the crossbars and result in 5.8x speedup comparing to 1x crossbar. The baseline fails to map the model when constrained to 80\% of crossbars.


\subsubsection{RRAM-aware NAS}
\begin{figure}[!t]
    \centering
    \includegraphics[width=0.4\textwidth]{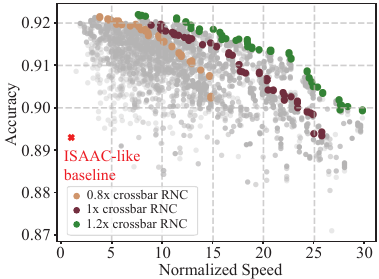}
    \vspace{-6pt}
    \caption{Pareto front and distribution of accuracy and speed of samples in RRAM-awared NAS with various amount of crossbar resource with baseline as red cross on chart.}
    
    \label{fig:pareto_front}
\vspace{-16pt}
\end{figure}

Compilation optimizes the execution of the network on RRAM crossbars without tailoring the network's architecture. Pairing that with RRAM-aware NAS further enhances overall system performance (measured as a combination of weighted accuray and speed). The results from the NAS are presented in Fig. \ref{fig:pareto_front}, which shows the Pareto front for different crossbar resources, compared against the ISAAC-like baseline. The Pareto front represents a set of optimal solutions, with the accuracy-oriented solutions positioned at the top and the speed-oriented solutions at the right. Notably, the Pareto front shows significant improvement as the number of crossbars increases from 0.8x to 1.2x.
The analytical result in Fig. \ref{fig:nas_result} demonstrates the advantages of NAS over both the baseline and compilation alone. With the 1x crossbar configuration as the baseline, the speed-optimized NAS solution achieves a 14.6x speedup. Meanwhile, the accuracy-optimized NAS solution achieves a 8.9x speedup while improving accuracy by a 3\% margin.

When resources are constrained to 0.8x the number of crossbars, the ISAAC-like mapping struggles to deploy all weight matrices onto the crossbars. In contrast, the compilation approach successfully deploys the network, resulting in similar accuracy and speed. The paired NAS and compilation shows further improvement, achieving both a 3\% increase in accuracy and a 13.8x speed improvement.
With additional crossbar resources (e.g. 1.2x crossbar), duplication of large layer boxes become feasible. This results in speed improvements for both accuracy-optimized and latency-optimized models, with gains of 7.6x and 29.8x, respectively. However, the accuracy of acc-optimized remains 0.92 regardless of the crossbar resources.

Overall, the paired RRAM-aware NAS and compilation consistently achieves over 80\% crossbar utilization, outperforming the ISAAC-like mapping by a margin of over 40\%. 

\section{Conclusion}
The RNC framework offers a comprehensive and robust solution that integrates RRAM-aware NAS and RRAM CIM compilation techniques to optimize DNNs for resource-limited edge devices. By leveraging advanced methods such as layer partitioning, weight duplication, and network packing, RNC significantly enhances the utilization of crossbar arrays while minimizing latency. The employment of a one-shot NAS with NSGA-II, guarantees the discovery of efficient and Pareto-optimal solutions. The experimental outcomes corroborate the effectiveness of the RNC framework, showcasing considerable performance enhancements and laying a solid foundation for the practical and efficient deployment of neural networks on edge CIM systems.

\section*{Acknowledgment}
This research is supported by the National Key R\&D Program of China (Grant No. 2023YFB2806300), National Natural Science Foundation of China (Grant Nos. 62122004, 62374181),  Beijing Natural Science Foundation (Grant No. Z210006), Hong Kong Research Grant Council (Grant Nos. 27206321,29517205922, 17212923). This research is also partially supported by ACCESS – AI Chip Center for Emerging Smart Systems, sponsored by Innovation and Technology Fund (ITF), Hong Kong SAR.


\bibliographystyle{unsrt}
\bibliography{ref}

\end{document}